\documentclass{article}
\usepackage{appendix}
\usepackage[a4paper, total={6in, 9in}]{geometry}
\usepackage{xurl}
\usepackage{float}
\usepackage{xcolor}
\usepackage[numbers]{natbib}
\usepackage{fancyvrb}
\usepackage{fvextra}
\usepackage{amsmath}
\usepackage{abstract}
\usepackage{lipsum}
\usepackage{subcaption}
\usepackage{graphics}
\usepackage[pipeTables=true]{markdown}
\usepackage{hyperref}
\usepackage{booktabs}

\title{ANLS* - A Universal Document Processing Metric for Generative Large Language Models}

\author{
    David¹ \\ Peer
    \and
    Philemon¹ \\ Schöpf
    \and
    Volckmar \\ Nebendahl
    \and
    Alexander \\ Rietzler
    \and
    Sebastian \\ Stabinger
}

\date{%
    DeepOpinion \\
    \url{https://deepopinion.ai} \\
    \vspace{0.8cm}
    \today
}

\newcommand{\type}[1]{\texttt{type}(#1)}
\DeclareMathOperator*{\argmax}{argmax}

\newcommand{\david}[1]{\emph{\textcolor{blue}{(David: #1)}}}
\newcommand{\philemon}[1]{\emph{\textcolor{purple}{(Philemon: #1)}}}
\newcommand{\sebastian}[1]{\emph{\textcolor{gray}{(Sebastian:  #1)}}}
\newcommand{\volckmar}[1]{\emph{\textcolor{orange}{(Volckmar: #1)}}}
\newcommand{\alex}[1]{\emph{\textcolor{red}{(Alex: #1)}}}

\renewcommand{\david}[1]{}
\renewcommand{\philemon}[1]{}
\renewcommand{\sebastian}[1]{}
\renewcommand{\volckmar}[1]{}
\renewcommand{\alex}[1]{}

\begin{document}
\maketitle

\def\thefootnote{1}\footnotetext{Equal contribution. Contact via \texttt{firstname.lastname@deepopinion.ai}}\def\thefootnote{\arabic{footnote}}

\def\thefootnote{2}\footnotetext{This paper was written with the assistance of GPT-4.}\def\thefootnote{\arabic{footnote}}

\def\ie{\emph{i.e}\onedot}

\begin{abstract}
Traditionally, discriminative models have been the predominant choice for tasks like document classification and information extraction. These models make predictions that fall into a limited number of predefined classes,  facilitating a binary true or false evaluation and enabling the direct calculation of metrics such as the F1 score. However, recent advancements in generative large language models (GLLMs) have prompted a shift in the field due to their enhanced zero-shot capabilities, which eliminate the need for a downstream dataset and computationally expensive fine-tuning. However, evaluating GLLMs presents a challenge as the binary true or false evaluation used for discriminative models is not applicable to the predictions made by GLLMs.

This paper introduces a new metric for generative models called ANLS* for evaluating a wide variety of tasks, including information extraction and classification tasks. The ANLS* metric extends existing ANLS metrics as a drop-in-replacement and is still compatible with previously reported ANLS scores. An evaluation of 7 different datasets, and more than 20 different GLLMs together with 3 different prompting methods using the ANLS* metric is also provided, demonstrating the importance of the proposed metric.

We also benchmark a novel approach to generate prompts for documents, called SFT, against other prompting techniques such as LATIN. In almost all cases, SFT outperforms other techniques and improves the state-of-the-art, sometimes by as much as $10$ percentage points.

Sources are available at \url{https://github.com/deepopinion/anls_star_metric}
\end{abstract}

\section{Introduction}

The increases in model size, dataset size, and available compute have significantly advanced the state-of-the-art (SOTA) on many natural language processing (NLP) tasks \citep{llm_scaling_law, bert, glp, lbe}. A unique and challenging domain within NLP is document processing, because documents contain text, images and tables and are, therefore, inherently multimodal. Additionally, the text is not strictly arranged in a linear fashion, but often has distinct positional structure within the 2D space of the document (i.e. the layout of the document). As such, document processing tasks are often addressed with specialized layout models \citep{layoutlm, layoutlmv2,layoutlmv3} that improve the performance by encoding important 2D positional information of bounding boxes together with the text tokens.

While discriminative models such as LayoutLMv3 \citep{layoutlmv3} have significantly advanced the state-of-the-art in document processing tasks, they still possess certain limitations. For instance, they are incapable of executing tasks that require additional synthesis, translation, or enhancement of text, as they cannot generate tokens and usually only label tokens. For example, a task may require extracting the date-time and transforming it into the \texttt{YYYY-MM-DD} format. Consequently, generative large language models (GLLMs) have garnered considerable attention in this field in recent times \citep{docllm} as they can solve these problems without the need for additional post-processing steps. Furthermore, these models are usually pre-trained on large datasets, eliminating the need for fine-tuning them on a specific downstream task as is usually done for classical deep learning models \citep{layoutlm,layoutlmv2,layoutlmv3}. Simply altering the input prompt (zero-shot) or providing a few examples demonstrating the task (few-shot) is sufficient to accomplish a task with decent performance. This is particularly crucial for document processing tasks, as the number of available datasets is limited and the creation of new ones is costly. Consequently, the community is moving towards large generative models for document processing tasks \citep{docllm, latin}.

While the evaluation of discriminative models typically relies on measuring the F1 score by counting correctly labeled bounding boxes coming from an OCR solution, this method is not applicable for GLLMs since the extracted, and possibly already pre-processed, information is directly returned as generated text. I.e. there is no direct connection between the OCR bounding boxes and the extracted information anymore. Generated text may also contain minor errors such as typos, which should be penalized differently compared to completely wrong answers. GLLMs are, therefore, typically evaluated using the Average Normalized Levenshtein Similarity (ANLS) metric \citep{anls}, which is a normalized form of the Levenshtein similarity, assessing the closeness of a generated answer to the expected output. A shortcoming of the ANLS metric is that it can only deal with strings and lists, but cannot be used for dictionaries or any combination of types that are often encountered when dealing with information extraction tasks. Additionally, some tasks require to extract information with a list structure such as line-item extraction \citep{line_item_dataset}, which requires the evaluation of complex output objects.

In this paper, we introduce ANLS*, a metric that can be used to evaluate a wide variety of tasks such as information extraction or classification, even in cases where output values that may contain minor errors are generated. It is worth mentioning, that this metric can also be used for discriminative models which allows for direct comparison of both discriminative and generative models with one single metric in the future. Additionally, the ANLS* metric can be applied to unstructured as well as structured outputs or any combination of both which makes it a versatile tool for the evaluation of document processing tasks. The proposed metric extends all previously defined ANLS metrics \citep{anls}. That is, results that could be calculated using the ANLS metric remain unchanged under ANLS*, while simultaneously offering greater flexibility. Therefore, it serves as a plug-in replacement for existing ANLS metrics.

Lastly, we provide qualitative as well as quantitative experiments using the proposed ANLS* metric to demonstrate the importance of the proposed metric. Various GLLMs and prompting methods across different datasets are evaluated with the proposed metric. Those results are not only provided for the community as a baseline for future experiments, but the scripts to reproduce the results are also publicly available.

\section{Related Work}

\paragraph{Metrics}
The normalized Levenshtein similarity (NLS) was defined by \citet{nls} to measure the similarity between words using the minimal distance between those words. Later, the Average Normalized Levenshtein Similarity (ANLS) was introduced by \citet{anls} for the evaluation of visual question-answering (VQA) tasks. The ANLS metric takes OCR errors into consideration, which makes it suitable for generative models. \citet{anlsl} further expanded the ANLS metric for the comparison of lists by using the Hungarian matching algorithm from \citet{hungarian_matching} to find the best match between the ground truth list and the predicted list. Finally, \citet{anls_null} extended the normalized Levenshtein similarity to account for predictions that should be null. This metric is not only useful for verifying the correct handling of unanswerable questions but also for penalizing hallucinations generated by GLLMs.

\paragraph{LLMs for Document Processing}
\citet{layoutlm} introduced a novel layout-aware language model to encode bounding box information as well as visual information about the document in tokens in order to improve document processing tasks. They outperformed purely text-based models such as BERT \citep{bert} by a large margin. Novel developments with different attention layers and pre-training methods have later been introduced \citep{layoutlmv2,layoutlmv3}. A novel GLLM called DocLLM was introduced by \citet{docllm} specifically for document processing tasks. This model captures cross-alignment between text and spatial modalities decomposing the attention mechanism of transformers. Although they could show significant improvements and novel pre-training methods, we will demonstrate that this model is still behind extremely large, purely text-based models, such as gpt-4 \citep{gpt4}. As a consequence, special prompting mechanisms that may encode OCR scanned documents in a way that is easier to understand by purely text-based models gained interest recently. \citet{latin} introduced such a prompting technique, called LATIN, to enhance the representation of documents for text-based GLLMs. Instead of directly encoding positional information in the tokens, they utilized layout-aware instruction prompts by using the positional information of the bounding boxes after the OCR scan. We developed a more advanced approach, called SFT that takes several properties of documents and LLMs into account. We will show that SFT is superior to LATIN and other prompting techniques. \footnote{The scope of this paper does not include an introduction of SFT. It may be introduced in a future publication.}

Other approaches completely bypass the conversion from OCR to prompts by employing a multimodal model that directly processes documents without a separate OCR step. For instance, \citet{ocr-free-doc-transformer} developed an OCR-free document transformer architecture. \citet{mplug-docowl} developed a generative multimodal model named MPlug-DocOWL that was trained on language-only, general vision-and-language, and document instruction tuning datasets.

\section{Metric Definition}
In this section, we introduce ANLS*. The goal is to develop a metric that is not only compatible with the existing ANLS \cite{anls} and the ANLSL \cite{anlsl} metrics, but also penalizes unanswerable questions in case they are answered as proposed by \citet{anls_null}. Additionally, the ANLS* metric should be a tool that is applicable for a wide variety of tasks, including tasks with dictionary outputs, lists or any combination of those in order to handle e.g. line-item extraction \citep{line_item_dataset} as well as simple question-answering tasks. As a result, the ANLS* metric serves as a direct drop-in replacement for all standard ANLS metrics defined by the community so far, and can additionally be used for evaluating all document-processing tasks.

\subsection{Supported data types}
The ANLS* metric supports the following data types:
\begin{enumerate}
    \item \texttt{String} - To compare strings against each other using the normalized Levenshtein similarity.
    \item \texttt{None} - Sometimes questions are not answerable. With this type it can be checked, whether the model does not answer. Any answer other than None will be penalized. On the other hand, if a model generates e.g. a None key in a dictionary that is not in the ground truth, ANLS* ignores it rather than penalizing or rewarding it.
    \item \texttt{Tuple} - Compare the given answer with each element in the tuple and select the element that produces the maximum ANLS* score. This is also provided by the classical ANLS metric \citep{anls}.
    
    \item \texttt{List} - Sometimes it is required to extract information in the form of lists from a document. For example, extracting all purchased items found in an invoice. While the order is not important, the list should contain all items. Note that the same item can occur multiple times in lists. Hungarian matching \citep{hungarian_matching} is used to compare the ground truth and the predicted list against each other. Both, missing elements as well as hallucinated elements, are penalized as introduced by \citet{anlsl}.
    \item \texttt{Dict} - For document information extraction it is usually required to extract key-value pairs. For example, when extracting the date and total value from an invoice. Missing keys as well as hallucinated keys are penalized.
\end{enumerate}

\begin{figure}
    \centering
    \begin{subfigure}{0.5\textwidth}
        \includegraphics[width=\textwidth]{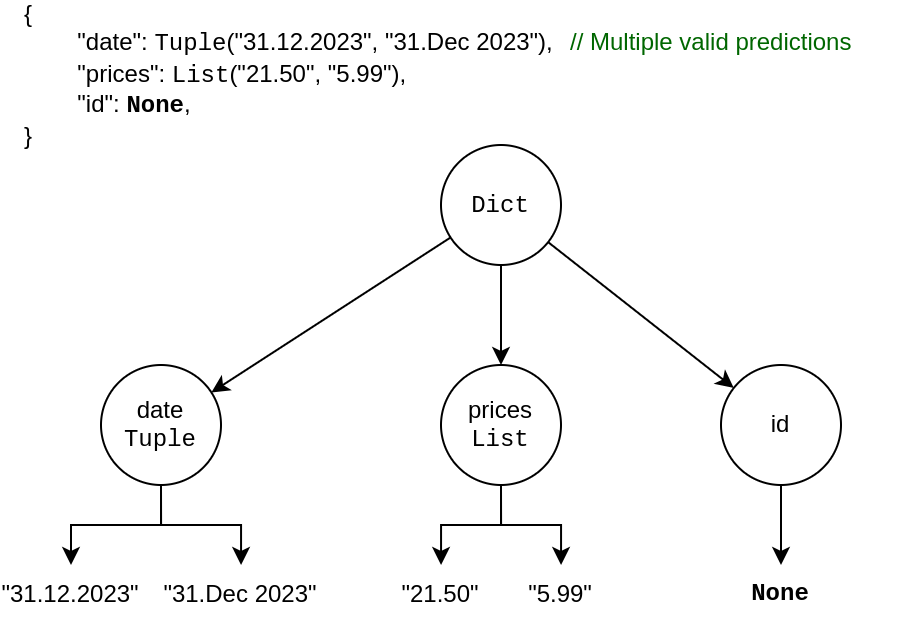}
        \caption{Ground truth.}
        \label{fig:ground_truth}
    \end{subfigure}

    \vspace*{0.8cm}

    \begin{subfigure}{0.35\textwidth}
        \includegraphics[width=\textwidth]{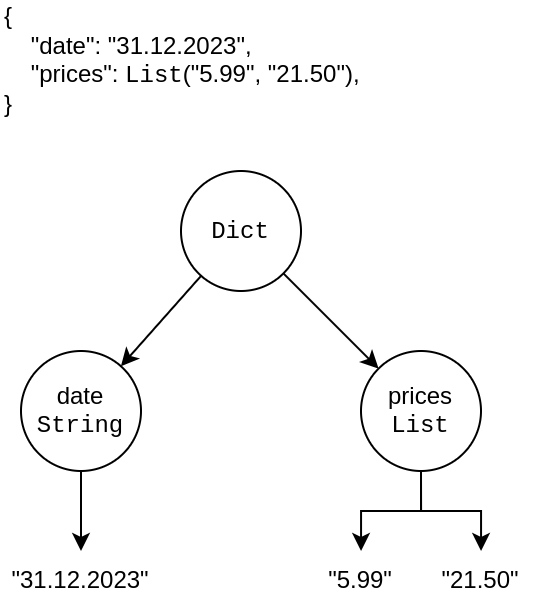}
        \caption{Prediction with $\text{ANLS*}=1.0$.}
        \label{fig:p_correct}
    \end{subfigure}
    \qquad \qquad \qquad
    \begin{subfigure}{0.41\textwidth}
        \includegraphics[width=\textwidth]{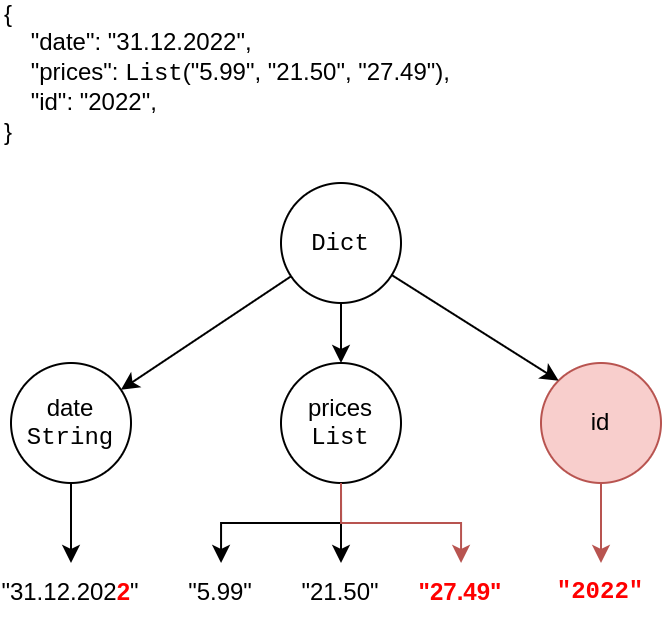}
        \caption{Prediction with $\text{ANLS*}<1.0$.}
        \label{fig:p_wrong}
    \end{subfigure}

    \caption{Examples of how the ground truth, as well as predictions, are decomposed into a tree structure. A correct prediction is shown in \autoref{fig:p_correct}, while \autoref{fig:p_wrong} visualizes a partially incorrect prediction. Its worth mentioning that any hallucination as well as incorrect types are penalized as well. More examples are given in \autoref{tbl:cases}.}
    \label{fig:figures}
\end{figure}

It is worth mentioning that all combinations of the above types are supported as well. For example, a dictionary may contain lists of strings or the elements of a list may be dictionaries. The implementation of the ANLS* metric maps those complex structures into a tree and compares the ground truth tree against the predicted tree from the model. \autoref{fig:ground_truth} visualizes how the ground truth is decomposed into a tree structure that can then be compared against predictions for an example.

\autoref{fig:p_correct} demonstrates a prediction with $\text{ANLS*} = 1.0$ w.r.t \autoref{fig:ground_truth}.
Finally, \autoref{fig:p_wrong} shows an example of a partially incorrect prediction. Note that the ANLS* metric is not only able to detect wrong strings but also wrong output structures. For example, the prediction may be a list, although a dictionary was expected. All these cases are correctly handled by the ANLS* metric.

\subsection{Formal definition of the ANLS* metric}
In the following, the ground truth is denoted as $g$ and the prediction as $p$.
Note that the type of the ground truth $\type{g}$ may differ from the type of the prediction $\type{p}$ in case the GLLM returns incorrect results. For example, the GLLM may answer with a sentence although a list was expected. The idea of the metric is to generate a tree from the ground truth as well as a tree from the prediction and to compute a matching between both trees. Additionally, the metric is normalized to account for different lengths of the ground truth and the prediction. Overall, the ANLS* metric is defined as follows:

\begin{align}
    \text{ANLS*}(g, p) = \frac{s(g, p)}{l(g, p)}
\end{align}

where $s$ is the score between the ground truth and the prediction and $l$ is the size of the trees $g$ and $p$ such that $\text{ANLS*}(g, p) \in [0, 1]$. Additionally, it is worth noting that each prediction in this tree is given equal weight. This implies that leaf nodes of large sub-trees carry the same weight as leaf nodes that appear at the top level.
\subsubsection{Definition of the score $s$}
The score $s$ is defined recursively to measure the similarity between the ground truth and the prediction. Note that in order to distinguish between the \emph{one of} semantic of a ground truth list used by the original ANLS metric, and the matching semantic implemented in the ANLSL metric, we introduced Tuples for the former and Lists for the latter. The score $s$ is defined as follows:
\begin{align}
    s(g, p) &= \begin{cases}
        1.0 & \text{if } \type{g} = \type{p} = \text{None} \\
        1.0 - \frac{\text{LD}(g, p)}{\max(|g|, |p|)} & \text{if } \type{g} = \type{p} = \text{String} \text{ and } s(g, p) \geq \tau \\
        s(g_i, p) \text{ with } i = \argmax_i(\text{ANLS*}(g_i, p)) & \text{if } \type{g} = \text{Tuple} \\
        \sum\limits_{(g_i, p_i) \in \psi(g, p)} s(g_i, p_i) & \text{if } \type{g} = \type{p} = \text{List} \\
        \sum\limits_{k \in \texttt{keys}(p)} s(g_k, p_k) & \text{if } \type{g} = \type{p} = \text{Dict} \text{ and } k \in \text{keys}(p) \\
        0.0 & \text{otherwise}
    \end{cases}
\end{align}
with $\text{LD}$ being the Levensthein distance and $\tau$ being the normalized Levensthein distance threshold which is set to $\tau = 0.5$. $\psi$ is the Hungarian matching algorithm \citep{hungarian_matching} performed according to the pairwise ANLS* of each ground truth and prediction element. This algorithm returns an optimal matching of elements between two lists w.r.t. a given score. The score for type mismatches (i.e., different subtrees) yields a score of $0.0$. The function $\texttt{keys}(x)$ returns all keys of a dictionary $x$, that are not None. It is important that $\texttt{keys}(x)$ ignores None values in order to penalize hallucinations correctly.

\subsubsection{Definition of the length $l$}
To normalize $s$, we define the length $l$ of each type as follows:
\begin{align}
    l(g, p) &= \begin{cases}
        1 & \text{if } \type{x} = \type{p} = \text{None} \\
        1 & \text{if } \type{g} = \type{p} = \text{String}\\
        l(g_i, p) \text{ with } i= \argmax_i(\text{ANLS*}(g_i, p)) & \text{if } \type{g} = \text{Tuple} \\
        \sum\limits_{(g_i, p_i)\in \psi(g, p)} l(g_i, p_i)  & \text{if } \type{g} = \type{p} = \text{List} \\ \qquad + \sum\limits_{g_u \notin \psi(g, p)} l_t(g_u) \\ \qquad + \sum\limits_{p_u \notin \psi(g, p)} l_t(p_u) \\
        \sum\limits_{k \in \texttt{keys}(p) \cap \texttt{keys}(q)} l(g_k, p_k) & \text{if } \type{g} = \type{p} = \text{Dict}  \\ \qquad + \sum\limits_{k \in \texttt{keys}(g) - \texttt{keys}(p)} l_t(g_k) \\ \qquad + \sum\limits_{k \in \texttt{keys}(p) - \texttt{keys}(g) } l_t(p_k) \\
        \max(l_t(g), l_t(p)) & \text{otherwise}
    \end{cases}
\end{align}
As can be seen, the length is weighted for all type matches accordingly. Nevertheless, it is not guaranteed that the prediction produced the correct output structure. On the other hand, a partially correct structure should not get a sore of $0.0$. To this end, we match the subtree and penalize wrong types via another length function $l_t$. It can be seen that the maximum length is used between the ground truth and the prediction - $\max(l_t(g), l_t(p))$ -- such that both, missing subtrees, as well as hallucinated subtrees, are penalized equally. The length function $l_t$ is defined as follows:

\begin{align}
    l_t(x) &= \begin{cases}
        1 & \text{if } \type{x} = \text{None} \\
        1 & \text{if } \type{x} = \text{String}\\
        \max_{x_i \in x}l_t(x_i) & \text{if } \type{x} = \text{Tuple} \\
        \sum\limits_{x_i \in x}l_t(x_i) & \text{if } \type{x} = \text{List} \\
        \sum\limits_{k \in \texttt{keys}(x)} l_t(x_k) & \text{if } \type{x} = \text{Dict} \\
    \end{cases}
\end{align}
where $x$ is either a (sub)tree of the prediction $p$ or a (sub)tree of the ground truth $g$.

Using the ANLS* metric we will next showcase some examples to demonstrate the behavior of the metric. In a quantitative study, we will later show that ANLS* is a suitable metric for a wide variety of tasks.

\section{Experimental Evaluation}
In this section, we evaluate the performance of the ANLS* metric both qualitatively and quantitatively. The source code required to reproduce these results can be accessed at \url{https://github.com/deepopinion/anls_star_metric}.

\subsection{Examples}
Different ANLS* scores for various ground truths and predictions are provided below, to offer the reader some insight into which predictions are considered as good and which are considered as bad. We also show some limitations of the proposed method. Note that tuples are interpreted as \emph{one of} and lists are interpreted as \emph{all of}.

\begin{table}[ht]
  \centering
    \caption{ANLS* scores for different predictions and ground truth types. \label{tbl:cases}}
    \resizebox{\columnwidth}{!}{
    \begin{tabular}{llllc}
    \toprule
        Id & Description & Ground Truth & Prediction & ANLS* \\
    \midrule
        1 & Correct String & \texttt{Hello World} & \texttt{Hello World} & 1.0 \\
        2 & Typo & \texttt{Hello World} & \texttt{Hello Wolrd} & 0.82 \\
        3 & Incorrect String & \texttt{Hello World} & \texttt{How are you?} & 0.0 \\
        4 & Hallucination & \textbf{None} & \texttt{Hello World!} & 0.0 \\
        5 & One of n & \texttt{tuple(Hello, World)} & \texttt{Hello} & 1.0 \\
        6 & Typo in one of n & \texttt{tuple(Hello, World)} & \texttt{Wolrd} & 0.6 \\
        7 & Expected String & \texttt{Hello World} & \texttt{list(Hello, World)} & 0.0 \\
        8 & Correct List & \texttt{list(Hello, World)} & \texttt{list(World, Hello)} & 1.0\\
        9 & Missing Element & \texttt{list(Hello, World)} & \texttt{list(Hello)} & 0.5 \\
        10 & Correct Dict & \texttt{\{a:Hello, b:World\}} & \texttt{\{b:World, a:Hello\}} & 1.0 \\
        11 & Missig Key & \texttt{\{a:Hello, b:World\}} & \texttt{\{a:Hello\}} & 0.5 \\
        12 & Hallucinated Key & \texttt{\{a:Hello, b:World\}} & \texttt{\{b:World, a:Hello, c:Great\}} & 0.67 \\
        13 & Complex Object & \texttt{\{a:Hello, b:list(W,r,l,d)\}} & \texttt{\{a:Hello, b:list(w,r,d)\}} & 0.8 \\
    \bottomrule
    \end{tabular}
    }
\end{table}

\begin{table}[ht]
  \centering
  \caption{ANLS* scores for edge cases.\label{tbl:edge_cases}}
    \resizebox{\columnwidth}{!}{
    \begin{tabular}{llllc}
        \toprule
        Id & Description & Ground Truth & Prediction & ANLS* \\
        \midrule
        14 & \texttt{list} casted implicitly to \texttt{tuple} & \texttt{list(Hello, World)} & \texttt{Hello} & 1.0 \\
        15 & Comparison of numbers & \texttt{0.2} & 0.199999999 & 0.0 \\
        16 & Incorrect Format & \texttt{31.12.2023} & \texttt{31.Dec 2023} & 0.58 \\
        17 & Unanswerable Question - Incorrect Answer & \texttt{Yesterday} & \texttt{Last Week} & 0.0 \\
        18 & Unanswerable Question - No Answer & \texttt{Yesterday} & \texttt{None} & 0.0 \\
        \bottomrule
    \end{tabular}
    }
\end{table}

Several cases for good and bad predictions including type mismatches are shown in \autoref{tbl:cases}. Additionally, we added some edge cases in \autoref{tbl:edge_cases} that may seem counter-intuitive at first, but they are required in order to keep the metric consistent with the previously defined ANLS and ANLSL metrics.

The first edge case \#14 shown in \autoref{tbl:edge_cases} (list automatically casted to a tuple) is implemented to ensure compatibility with common datasets where possible answers are returned as lists, while the actual answer is a single string. According to the proposed semantics, all possible answers should be tuples. However, to ensure the reproducibility of experiments with classical QA datasets, we implemented this implicit casting in cases where the ground truth is a list and the prediction is a string. In case 15, it is evident that numbers are not interpreted as numbers, but a string comparison is performed instead. Case 16 illustrates that different formats may produce a high error, even though the semantics are the same. Lastly, cases 17 and 18 demonstrate that completely incorrect answers are weighted equally to missing answers.

\subsection{Experimental Evaluation}
In this subsection, the quantitative evaluation of the ANLS* metric on many different datasets and different GLLMs is shown. More precisely, we evaluated two QA datasets (DocVQA \citep{doc_vqa}, MPDocVQA \citep{mp_doc_vqa}) and five information extraction datasets (Kleister Charity \citep{kleister}, Kleister NDA \citep{kleister}, SROIE \citep{sroie}, VRDU Ad Buy \citep{vrdu}, VRDU Registration\citep{vrdu}). 
We used LangChain \citep{langchain} for the implementation. Whenever a request to a provider failed (e.g. network timeouts etc.), we send the same request with a sleep time of 10 seconds again. After 5 retries we considered the request as failed resulting in ANLS* of 0.0.  We used the following prompts to query each model:

\newpage
\paragraph{QA prompt}
\begin{Verbatim}[frame=single,breaklines=true]
You are a world-class question answering system.
You are given a document and a question. You must answer the question based on the document.
Precisely answer the question without any additional text.
Its very important to NOT write full sentences!
Note: Ensure that the answer is precisely contained in the original document.
Here is the document:
{document}
Here is the question:
{question}
\end{Verbatim}

\paragraph{Information extraction prompt}
\begin{Verbatim}[frame=single,breaklines=true]
You are a document information extraction system.
You are given a document and a json with keys that must be extracted from the document.
Here is the document:
{document}
{format_instructions}
\end{Verbatim}
Format extractions are automatically generated by LangChain according to the given dataset and keys that must be extracted. For representing the document itself, different prompting methods were evaluated (Simple, LATIN, SFT). More details can be found in the provided source code. For vision based models we excluded the document and added one user message per page directly including the corresponding image without executing any OCR.

Note that special prompting techniques were required for mistral-large as well as claude-3 in order to get decent values. We refer to the source code for more details. The results are shown in \autoref{tbl:quantitative_evaluation}.
More detailed results with intermediate versions of each model are given in the appendix.
For gemini-pro we set the safety filter to "High" to not measure false positives of the safety filter.

\begin{table}[ht]
     \caption{ANLS* score for different GLLMs and Datasets. For each model the latest version is reported. Scores for older versions can be found in the appendix. \\\tiny{No document prompting method is required for vision-based models. Therefore Latin Prompting and SFT is not executed for those.}\label{tbl:quantitative_evaluation}}

    \resizebox{\columnwidth}{!}{
        \begin{tabular}{llcccccccc}
            \hline
            \toprule    
            Dataset             & Method                & gpt-4.1         & gpt-4.5          & gemini-2.5-pro     & mistral-large  & claude-37     & Llama-3.1-405B    \\ 
            \midrule
            DocVQA              & Simple                & 0.678           & 0.760            & 0.755              & 0.445          & 0.763             &  0.653   \\
            ~                   & Latin Prompting       & 0.727           & 0.794            & 0.803              & 0.447          & 0.823             &  0.595   \\ 
            ~                   & \textbf{SFT (Ours)}   & 0.746           & 0.855            & \textbf{0.863}     & 0.648          & 0.834             &  0.737   \\  \addlinespace
            MPDocVQA            & Simple                & 0.671           & 0.677            & 0.744              & 0.364          & 0.713             &  0.625   \\ 
            ~                   & Latin Prompting       & 0.738           & 0.767            & 0.826              & 0.335          & 0.780             &  0.640   \\ 
            ~                   & \textbf{SFT (Ours)}   & 0.762           & 0.777            & \textbf{0.841}     & 0.476          & 0.790             &  0.753   \\ \addlinespace
            Kleister Charity    & Simple                & 0.776           & 0.765            & 0.755              & 0.652          & 0.778             &  0.665   \\
            ~                   & Latin Prompting       & 0.773           & 0.775            & 0.767              & 0.576          & 0.778             &  0.643   \\ 
            ~                   & \textbf{SFT (Ours)}   & 0.764           & 0.775            & 0.765              & 0.657          & \textbf{0.786}    &  0.680   \\ \addlinespace
            Kleister NDA        & Simple                & 0.602           & 0.638            & 0.594              & 0.637          & 0.629             &  0.608   \\
            ~                   & Latin Prompting       & 0.606           & 0.637            & 0.605              & 0.624          & 0.616             &  0.591   \\ 
            ~                   & \textbf{SFT (Ours)}   & 0.607           & \textbf{0.641}   & 0.610              & \textbf{0.641} & 0.64              &  0.610   \\ \addlinespace
            SROIE               & Simple                & 0.906           & 0.855            & 0.696              & 0.855          & \textbf{0.935}    &  0.853   \\
            ~                   & Latin Prompting       & 0.914           & 0.836            & 0.705              & 0.863          & 0.877             &  0.779   \\
            ~                   & \textbf{SFT (Ours)}   & 0.900           & 0.877            & 0.717              & 0.905          & 0.899             &  0.879   \\ \addlinespace
            VRDU AD Buy         & Simple                & 0.672           & 0.668            & 0.708              & 0.386          & 0.625             &  0.453   \\
            ~                   & Latin Prompting       & 0.725           & 0.725            & 0.742              & 0.435          & 0.639             &  0.546   \\ 
            ~                   & \textbf{SFT (Ours)}   & \textbf{0.754}  & 0.726            & 0.737              & 0.594          & 0.641             &  0.700   \\ \addlinespace
            VRDU Registration   & Simple                & \textbf{0.703}  & 0.664            & 0.597              & 0.579          & 0.658             &  0.614   \\
            ~                   & Latin Prompting       & 0.710           & 0.667            & 0.601              & 0.587          & 0.669             &  0.632   \\ 
            ~                   & \textbf{SFT (Ours)}   & 0.664           & 0.666            & 0.593              & 0.639          & 0.673             &  0.614   \\ \addlinespace
            \bottomrule
        \end{tabular}
    }
\end{table}

For the evaluation, we always report the values of the newest model version. A detailed comparison between different intermediate versions of 
GPT as well as Gemini is given in the appendix.

\paragraph{GLLM Comparison}
First of all, it can be seen that the only model that the best performing models are currently claude-3 and gemini-1.5-pro. 
Both models are good in both, VQA as well as IE tasks. So its not clear yet when to select one over the other. 
Nevertheless, it can be seen that gpt-4-turbo, gpt-4-vision as well as mistral-large are quite behind those 
two models. One interesting finding worth mentioning is that gemini-1.5-pro has a quite a restrictive safety filter 
which explains lower scores on the SROIE dataset as we found that this filter had quite some false-positives on the address field
of this dataset. So in future we expect that the same model with an improved safety filter will outperform the other models on even more datasets.
A detailed comparison of intermediate versions of GPT as well as Gemini is given in the appendix.

\paragraph{Different Prompting Methods}
\autoref{tbl:quantitative_evaluation} shows the result for three different prompting methods. 
The highest scores of 5 out of 7 datasets were achieved with our SFT method. 
Only for Kleister Charity and Kleister NDA this was not the case. Results for KleisterNDA 
are very close. For Kleister Charity we found that documents
 contain mainly text and almost no positional encoding which may describe why SFT 
 and LATIN did not improve simple prompting. Nevertheless, documents usually contain 
 2D structures and not purely text which shows the importance of specialized prompt formatting 
 techniques such as SFT.

\paragraph{GLLMs with Vision Capabilities}
Prompting techniques are mainly required as those models only support text input. Nevertheless, 
some models exist that support vision capabilities. Although the gpt-4-vision model reaches 
decent scores, it is still behind our SFT solution which exploits an OCR scanner. In all cases 
gpt-4-turbo + SFT outperformed gpt-4-vision. This is a strong indicator that vision models are 
not yet competitive against specialized OCR scanners when combined with advanced prompting techniques.

\paragraph{Comparison with DocLLM}
\citet{docllm} developed and trained a special generative model for documents with 7B parameters. 
They report 0.634 on DocVQA and 0.499 zero-shot scores on Kleister Charity. Those scores are quite 
behind our SFT solution with 0.831 on DocVQA and 0.800 on Kleister Charity. This shows that smaller 
specialized models are not yet competitive against large purely text-based models that exploit 
specialized document prompting techniques.

\paragraph{Open-source Models}
We also made tests with mixtral-8x7b, but found that those models are still far behind closed-source 
models. We, therefore, decided to not include those models in the final evaluation and will report 
additional results in a future or updated publication.

\section{Conclusion \& Discussion}
In this paper, we introduce a novel metric called ANLS*, which serves as a plug-in replacement for existing ANLS metrics. This metric is not only applicable for traditional GLLM tasks such as QA, but also for information extraction tasks, and even for more complex outputs. We demonstrate that ANLS* is a versatile metric for a broad range of tasks. We evaluate the ANLS* metric using three different GLLMs on seven different datasets. We also demonstrate that more advanced methods such as SFT outperform prompting techniques, such as LATIN. We found that claude-3 is competitive against gpt-4-turbo, but all others are still behind those two models. For DocLLM we beliefe that this specially trained model is still too small with 7B parameters to be competitive against larger pure text-based models. Unfortunately, it is not available at the time of writing such that combinations of prompting techniques with DocLLM can not be tested yet.

We posit that ANLS* is a suitable metric for the evaluation of generative models and should be adopted for future use. We also claim that the ANLS* metric is suitable for discriminative models, allowing for a comparison of generative and discriminative models using a single metric. We hope that the ANLS* metric will be adopted by the community in the future.

\nocite{*}
\bibliographystyle{plainnat}
\bibliography{refs}

\newpage
\appendixpage
\appendix
\section{Results GPT}
In this appendix, a detailed evaluation of different versions of GPT is shown. 

\begin{table}[H]
    \caption{ANLS* score for different versions of GPT. \\\tiny{No document prompting method is required for vision-based models. Therefore Latin Prompting and SFT is not executed for those.}\label{tbl:quantitative_evaluation_openai}}

   \resizebox{\columnwidth}{!}{
       \begin{tabular}{llccccccccccccc}
           \hline
           \toprule    
           Dataset             & Method                & gpt-3.5-turbo-16k      & gpt-3.5-turbo         & gpt-4                 & gpt-4-turbo    & gpt-4-vision & gpt-4o            & gpt-4o + vision & gpt-4o-mini & gpt-4o        & gpt-4.5	         & gpt-4.1      & gpt-4.1-mini & gpt-4.1-nano \\ 
                               & Version               & 0613                   & 0125                  & 1106-preview          & 2024-04-09     & preview-1106 & 2024-05-13        & 2024-05-13      & 2024-07-18  & 2024-08-06    & preview-2025-02-27 & 2025-04-14   & 2025-04-14   & 2025-04-14 \\
           \midrule
           DocVQA              & Simple                & 0.576                  & 0.550                 & 0.607                 & 0.589          & 0.759        & 0.656             & -               & 0.568       &  0.72         & 0.760              & 0.678        & 0.616        & 0.640 \\
           ~                   & Latin Prompting       & 0.620                  & 0.626                 & 0.699                 & 0.631          & -            & 0.718             & -               & 0.623       &  0.742        & 0.794              & 0.727        & 0.672        & 0.713 \\ 
           ~                   & \textbf{SFT (Ours)}   & 0.768                  & 0.733                 & 0.790                 & 0.770          & -            & 0.747             & 0.754           & 0.754       &  0.809        & 0.855              & 0.746        & 0.750        & 0.760 \\  \addlinespace
           MPDocVQA            & Simple                & 0.507                  & 0.461                 & 0.635                 & 0.606          & 0.708        & 0.686             & -               & 0.577       &  0.623        & 0.677              & 0.671        & 0.584        & 0.632 \\ 
           ~                   & Latin Prompting       & 0.489                  & 0.507                 & 0.739                 & 0.656          & -            & 0.718             & -               & 0.584       &  0.736        & 0.767              & 0.738        & 0.629        & 0.707 \\ 
           ~                   & \textbf{SFT (Ours)}   & 0.724                  & 0.731                 & 0.781                 & 0.760          & -            & 0.762             & 0.789           & 0.731       &  0.735        & 0.777              & 0.762        & 0.691        & 0.764 \\ \addlinespace
           Kleister Charity    & Simple                & 0.527                  & 0.564                 & 0.743                 & 0.708          & 0.751        & 0.711             & -               & 0.647       &  0.723        & 0.765              & 0.776        & 0.755        & 0.649 \\
           ~                   & Latin Prompting       & 0.477                  & 0.493                 & 0.735                 & 0.728          & -            & 0.721             & -               & 0.672       &  0.736        & 0.775              & 0.773        & 0.770        & 0.639 \\ 
           ~                   & \textbf{SFT (Ours)}   & 0.534                  & 0.548                 & 0.763                 & 0.733          & -            & 0.737             & 0.749           & 0.707       &  0.752        & 0.775              & 0.764        & 0.765        & 0.683 \\ \addlinespace
           Kleister NDA        & Simple                & 0.361                  & 0.575                 & 0.695                 & 0.633          & 0.664        & 0.649             & -               & 0.622       &  0.626        & 0.638              & 0.602        & 0.623        & 0.595 \\
           ~                   & Latin Prompting       & 0.412                  & 0.593                 & 0.705                 & 0.643          & -            & 0.646             & -               & 0.633       &  0.635        & 0.637              & 0.606        & 0.626        & 0.579 \\ 
           ~                   & \textbf{SFT (Ours)}   & 0.315                  & 0.562                 & 0.703                 & 0.659          & -            & 0.655             & 0.648           & 0.628       &  0.625        & 0.641              & 0.607        & 0.624        & 0.586 \\ \addlinespace
           SROIE               & Simple                & 0.509                  & 0.706                 & 0.835                 & 0.849          & 0.834        & 0.887             & -               & 0.863       &  0.905        & 0.855              & 0.906        & 0.856        & 0.804 \\
           ~                   & Latin Prompting       & 0.723                  & 0.708                 & 0.851                 & 0.847          & -            & 0.863             & -               & 0.836       &  0.912        & 0.836              & 0.914        & 0.867        & 0.817 \\
           ~                   & \textbf{SFT (Ours)}   & 0.792                  & 0.649                 & 0.873                 & 0.890          & -            & 0.897             & 0.892           & 0.872       &  0.897        & 0.877              & 0.900        & 0.879        & 0.867 \\ \addlinespace
           VRDU AD Buy         & Simple                & 0.414                  & 0.084                 & 0.553                 & 0.483          & 0.640        & 0.596             & -               & 0.398       &  0.645        & 0.668              & 0.672        & 0.651        & 0.548 \\
           ~                   & Latin Prompting       & 0.424                  & 0.211                 & 0.586                 & 0.506          & -            & 0.655             & -               & 0.436       &  0.721        & 0.725              & 0.725        & 0.675        & 0.581 \\ 
           ~                   & \textbf{SFT (Ours)}   & 0.676                  & 0.452                 & 0.770                 & 0.712          & -            & 0.753             & 0.755           & 0.693       &  0.768        & 0.726              & 0.754        & 0.748        & 0.647 \\ \addlinespace
           VRDU Registration   & Simple                & 0.586                  & 0.615                 & 0.676                 & 0.609          & 0.665        & 0.605             & -               & 0.594       &  0.628        & 0.664              & 0.703        & 0.681        & 0.562 \\
           ~                   & Latin Prompting       & 0.606                  & 0.630                 & 0.673                 & 0.614          & -            & 0.611             & -               & 0.608       &  0.628        & 0.667              & 0.710        & 0.682        & 0.583 \\ 
           ~                   & \textbf{SFT (Ours)}   & 0.637                  & 0.654                 & 0.711                 & 0.590          & -            & 0.605             & 0.615           & 0.596       &  0.624        & 0.666              & 0.664        & 0.646        & 0.587 \\ \addlinespace
           \bottomrule
       \end{tabular}
   }
\end{table}

\newpage
\section{Results Gemini}
In this appendix, a detailed evaluation of different versions of Gemini is shown. 

\begin{table}[H]
    \caption{ANLS* score for different versions of Gemini. \\\tiny{No document prompting method is required for vision-based models. Therefore Latin Prompting and SFT is not executed for those.}\label{tbl:quantitative_evaluation_gemini}}

   \resizebox{\columnwidth}{!}{
       \begin{tabular}{llccccccccc}
           \hline
           \toprule    
           Dataset             & Method                & gemini-1.0-pro     & gemini-1.5-pro-preview        & gemini-1.5-pro    & gemini-1.5-pro + vision   & gemini-1.5-flash & gemini-2.0-flash & gemini-2.5-pro  & gemini-2.5-flash \\ 
                               & Version               & 001                & 0409                          & 001               & 001                       & 001              & 001              & preview-03-25   & preview-04-17\\
           \midrule
           DocVQA              & Simple                & 0.586              & 0.647                         & 0.655             & -                         & 0.64             & 0.689            & 0.755           & 0.750 \\
           ~                   & Latin Prompting       & 0.676              & 0.668                         & 0.734             & -                         & 0.707            & 0.773            & 0.803           & 0.826 \\ 
           ~                   & \textbf{SFT (Ours)}   & 0.741              & 0.737                         & 0.752             & 0.850                     & 0.821            & 0.770            & 0.863           & 0.814 \\  \addlinespace
           MPDocVQA            & Simple                & 0.603              & 0.627                         & 0.643             & -                         & 0.696            & 0.680            & 0.744           & 0.684 \\ 
           ~                   & Latin Prompting       & 0.502              & 0.726                         & 0.755             & -                         & 0.665            & 0.680            & 0.826           & 0.809 \\ 
           ~                   & \textbf{SFT (Ours)}   & 0.616              & 0.794                         & 0.794             & 0.870                     & 0.786            & 0.816            & 0.841           & 0.817 \\ \addlinespace
           Kleister Charity    & Simple                & 0.583              & 0.764                         & 0.748             & -                         & 0.720            & 0.735            & 0.755           & 0.759 \\
           ~                   & Latin Prompting       & 0.478              & 0.763                         & 0.747             & -                         & 0.732            & 0.753            & 0.767           & 0.768 \\ 
           ~                   & \textbf{SFT (Ours)}   & 0.633              & 0.779                         & 0.764             & 0.695                     & 0.752            & 0.754            & 0.765           & 0.775 \\ \addlinespace
           Kleister NDA        & Simple                & 0.623              & 0.722                         & 0.613             & -                         & 0.592            & 0.622            & 0.594           & 0.621 \\
           ~                   & Latin Prompting       & 0.599              & 0.707                         & 0.608             & -                         & 0.596            & 0.623            & 0.605           & 0.605 \\ 
           ~                   & \textbf{SFT (Ours)}   & 0.552              & 0.715                         & 0.623             & 0.520                     & 0.600            & 0.620            & 0.610           & 0.612 \\ \addlinespace
           SROIE               & Simple                & 0.263              & 0.594                         & 0.686             & -                         & 0.689            & 0.737            & 0.696           & 0.696 \\
           ~                   & Latin Prompting       & 0.371              & 0.623                         & 0.672             & -                         & 0.664            & 0.658            & 0.705           & 0.671 \\
           ~                   & \textbf{SFT (Ours)}   & 0.288              & 0.623                         & 0.681             & 0.931                     & 0.699            & 0.683            & 0.717           & 0.684 \\ \addlinespace
           VRDU AD Buy         & Simple                & 0.510              & 0.645                         & 0.545             & -                         & 0.597            & 0.674            & 0.708           & 0.651 \\
           ~                   & Latin Prompting       & 0.556              & 0.686                         & 0.579             & -                         & 0.628            & 0.706            & 0.742           & 0.682 \\ 
           ~                   & \textbf{SFT (Ours)}   & 0.685              & 0.765                         & 0.734             & 0.545                     & 0.701            & 0.746            & 0.737           & 0.717 \\ \addlinespace
           VRDU Registration   & Simple                & 0.699              & 0.730                         & 0.549             & -                         & 0.593            & 0.650            & 0.597           & 0.620 \\
           ~                   & Latin Prompting       & 0.740              & 0.749                         & 0.555             & -                         & 0.633            & 0.620            & 0.601           & 0.592 \\ 
           ~                   & \textbf{SFT (Ours)}   & 0.720              & 0.780                         & 0.585             & 0.570                     & 0.667            & 0.613            & 0.593           & 0.577 \\ \addlinespace
           \bottomrule
       \end{tabular}
   }
\end{table}

\newpage
\section{Results Claude}
In this appendix, a detailed evaluation of different versions of Claude is shown. 

\begin{table}[H]
    \caption{ANLS* score for different versions of Claude. \label{tbl:quantitative_evaluation_claude}}

   \resizebox{\columnwidth}{!}{
       \begin{tabular}{llcccccc}
           \hline
           \toprule    
           Dataset             & Method                & claude-3-opus      & claude-3-5-sonnet & claude-3-7-sonnet \\ 
                               & Version               & 20240229           & 20240620          & 20250219  \\  
           \midrule
           DocVQA              & Simple                & 0.768              &  0.653            & 0.763 \\
           ~                   & Latin Prompting       & 0.762              &  0.739            & 0.823 \\ 
           ~                   & \textbf{SFT (Ours)}   & 0.831              &  0.769            & 0.834 \\  \addlinespace
           MPDocVQA            & Simple                & 0.636              &  0.581            & 0.713 \\ 
           ~                   & Latin Prompting       & 0.438              &  0.561            & 0.780 \\ 
           ~                   & \textbf{SFT (Ours)}   & 0.575              &  0.582            & 0.790 \\ \addlinespace
           Kleister Charity    & Simple                & 0.800              &  0.774            & 0.778 \\
           ~                   & Latin Prompting       & 0.787              &  0.786            & 0.778 \\ 
           ~                   & \textbf{SFT (Ours)}   & 0.786              &  0.773            & 0.786 \\ \addlinespace
           Kleister NDA        & Simple                & 0.673              &  0.623            & 0.629 \\
           ~                   & Latin Prompting       & 0.670              &  0.626            & 0.616 \\ 
           ~                   & \textbf{SFT (Ours)}   & 0.677              &  0.635            & 0.640 \\ \addlinespace
           SROIE               & Simple                & 0.933              &  0.922            & 0.935 \\
           ~                   & Latin Prompting       & 0.926              &  0.89             & 0.877 \\
           ~                   & \textbf{SFT (Ours)}   & 0.949              &  0.917            & 0.899 \\ \addlinespace
           VRDU AD Buy         & Simple                & 0.577              &  0.569            & 0.625 \\
           ~                   & Latin Prompting       & 0.608              &  0.609            & 0.639 \\ 
           ~                   & \textbf{SFT (Ours)}   & 0.633              &  0.612            & 0.641 \\ \addlinespace
           VRDU Registration   & Simple                & 0.685              &  0.663            & 0.658 \\
           ~                   & Latin Prompting       & 0.715              &  0.664            & 0.669 \\ 
           ~                   & \textbf{SFT (Ours)}   & 0.705              &  0.670            & 0.673 \\ \addlinespace
           \bottomrule
       \end{tabular}
   }
\end{table}

\end{document}